\documentclass[journal]{IEEEtran}
\usepackage{hyperref}
\usepackage{cleveref}
\usepackage{calc}
\usepackage{algpseudocode}
\usepackage{algorithm}
\usepackage{mathtools}
\usepackage{amsmath}
\usepackage{cleveref}
\usepackage{calc}
\usepackage{amssymb}
\usepackage{easybmat}
\usepackage{amsthm}
  \newtheoremstyle{dotless}{}{}{\itshape}{}{\bfseries}{}{ }{}
  \theoremstyle{dotless}

\ifCLASSINFOpdf
\else
\fi
\begin{document}
%
\title{Multi-Task Kernel Null-Space for One-Class Classification}
%
%
%
\author{Shervin~Rahimzadeh~Arashloo and~Josef~Kittler,~\IEEEmembership{Life~Member,~IEEE}
\thanks{S.R. Arashloo. is with the Department of Computer Engineering, Bilkent University, Ankara, Turkey, e-mail: S.rahimzadeh@cs.bilkent.edu.tr}
\thanks{J. Kittler is with CVSSP, University of Surrey, Guildford, Surrey, UK}
\thanks{Manuscript received ? ?, 2019; revised ? ?, 2019.}}
%
%
\markboth{Journal of \LaTeX\ Class Files,~Vol.~?, No.~?, ?~2019}%
{Shell \MakeLowercase{\textit{et al.}}: Bare Demo of IEEEtran.cls for IEEE Journals}
%
\maketitle
\begin{abstract}
The one-class kernel spectral regression (OC-KSR), the regression-based formulation of the kernel null-space approach has been found to be an effective Fisher criterion-based methodology for one-class classification (OCC), achieving state-of-the-art performance in one-class classification while providing relatively high robustness against data corruption. This work extends the OC-KSR methodology to a multi-task setting where multiple one-class problems share information for improved performance. By viewing the multi-task structure learning problem as one of compositional function learning, first, the OC-KSR method is extended to learn multiple tasks' structure \textit{linearly} by posing it as an instantiation of the separable kernel learning problem in a vector-valued reproducing kernel Hilbert space where an output kernel encodes tasks' structure while another kernel captures input similarities. Next, a non-linear structure learning mechanism is proposed which captures multiple tasks' relationships \textit{non-linearly} via an output kernel. The non-linear structure learning method is then extended to a sparse setting where different tasks compete in an output composition mechanism, leading to a sparse non-linear structure among multiple problems. Through extensive experiments on different data sets, the merits of the proposed multi-task kernel null-space techniques are verified against the baseline as well as other existing multi-task one-class learning techniques.
\end{abstract}
%
\IEEEpeerreviewmaketitle
\section{Introduction}
%
%
%
%
\IEEEPARstart{I}{n} the presence of large within-class variations, pattern classification techniques typically require a sufficiently large and representative set of training data to provide reasonable generalisation performance. With the growing complexity of the learning problems and the associated decision-making systems, the need for larger sets of training data is accentuated even further. While there exist applications where data is abundantly available, there are other situations where the number of training observations relative to the complexity of the learning machine may not be readily increased. Such situations arise when the classification system is quite complex whereas the cost of collecting training samples is relatively high or samples are rare by nature. In other cases where sufficient training observations are accessible, for effective training, multiple passes through the available samples may be required, increasing the complexity of the learning stage. The problem is also manifest in the settings where a large number of training observations might exist but they fail to capture the real distribution of the underlying phenomena. Degeneration of training data when combined with the limitations of learning systems may lead to a sub-optimal performance in certain problems. Although other alternatives exist, in these situations, sharing knowledge among multiple tasks via the multi-task learning (MTL) paradigm has been found to be an effective strategy to improve performance when individual problems are in some sense related \cite{DBLP:journals/corr/ZhangY17aa}. Sharing knowledge among multiple problems may enhance the generalisation performance of individual learners, reduce the required number of training samples or the number of learning cycles needed to achieve a particular performance level by exploiting commonalities/differences among different problems. As such, MTL is known to be an effective mechanism to inductive transfer which enhances generalisation by exploiting the domain information available in the training signals of individual problems as an inductive bias \cite{Caruana1997}. This objective is typically achieved by learning multiple tasks in parallel while using a shared representation. 

Although other strategies exit, the MTL approach may be cast within the framework of the reproducing kernel Hilbert space for vector-valued functions (RKHSvv) \cite{Micchelli2004}. In this context, the problem may be viewed as learning vector-valued functions where each vector component is a real-valued function corresponding to a particular task. In the RKHSvv, the relationship between multiple inputs and the outputs is modelled by means of a positive definite multi-task kernel \cite{Evgeniou05learningmultiple}. A plausible and computationally attractive simplification of this methodology is offered by the separable kernel learning strategy, assuming a multiplicative decomposition of a multi-task kernel in terms of a kernel on the inputs and another on task indices \cite{Evgeniou05learningmultiple,Caponnetto:2008:UMK:1390681.1442785,10.1007/978-3-642-15880-3_10}. In this formalism, the input and outputs are decoupled in the sense that the input feature space does not vary by task and the structure of different tasks is solely represented through the corresponding output kernel. Since a decomposition of the multi-task kernel as a product of a kernel on the input and another on the output facilitates the optimisation of the kernel on the task indices simultaneously with learning the predictive vector-valued function, it is widely applied as a kernel-based approach to model learning problems with multiple outputs. 

The MTL strategy has been successfully applied in a variety of different problems \cite{}. A relatively challenging classification problem, among others, is known to be one-class classification (OCC) \cite{Chandola:2009:ADS:1541880.1541882}. OCC is defined as the problem of identifying patterns which conform to a specific behaviour (known as normal/target patterns) and distinguishing them from all other observations (referred to as novelties/anomalies, etc.). The interest in one-class learning is fuelled, in part, by the observation that very often a closed form definition of normality does exist whereas typically no such definition for an anomalous condition is available. While one-class classification forms the backbone of a wide variety of applications \cite{6846360,BEGHI20141953,4694106,Kamaruddin:2016:CCF:2980258.2980319,7435726,6566012,4668357}, it usually suffers from a lack of representative training samples. The complexity of the problem is mainly due to the difficulty of obtaining non-target samples for training or their propensity to appear in unpredictable novel forms during the operational phase of the system. These adversities associated with the OCC problem make it a suitable candidate to benefit from a multi-task learning strategy. While there exist some previous effort on utilising tasks' structures in designing one-class classification methods \cite{HE2014416,7899861,5596881,8215490}, they typically rely on different flavours of the support vector machines paradigm. A plausible alternative to the SVM formulation is that of regularised regression \cite{DBLP:conf/cikm/DangCWZC17}. The major challenges in the context of multi-target regression are known to be due to jointly modelling inter-target correlations and non-linear input-output relationships \cite{7888599}. By exploring the shared knowledge across relevant targets to capture the inter-target correlations in a non-OCC setting, the performance of multi-target regression has been shown to improve \cite{ICML2011Dinuzzo_54,Ciliberto:2015:CLM:3045118.3045283,7888599,7879814,JMLR:v17:15-602}. In practice, however, multiple outputs represent higher level concepts which generate highly complex relationships, that call for powerful non-linear regression models, commonly formulated in the reproducing kernel Hilbert space. However, even in a general context beyond the OCC paradigm, despite relying on a Hilbert space formulation, very often the relationship among multiple tasks in the existing multi-task regression methods is captured in a \textit{linear} fashion and represented in terms of an output mixing matrix, which limits the representational capacity of such methods.

In the current study, the kernel null-space technique for one-class classification \cite{6619277,8099922,7045967}, and in particular, its regression-based formulation known as one-class kernel spectral regression (OC-KSR) \cite{DBLP:journals/corr/abs-1807-01085,DBLP:journals/corr/abs-1902-02208} is extended to a multi-task learning framework. The OC-KSR method as compared with other alternatives has been found to provide better performance and computational efficiency while being more resilient against data corruption. In the context of the OC-KSR method, we illustrate that the relationship among multiple related OCC problems may be captured effectively by learning related tasks concurrently based on the notion of separability of the multi-task kernel. To this end, multiple one-class learning problems are modelled as the components of a vector-valued function while learning their structure corresponds to the choice of suitable functional spaces.
\subsection{Overview of the proposed approach}
As noted earlier, in this work, the kernel regression-based formulation of the Fisher null-space technique for the one-class classification problem is extended to benefit from a multi-task learning strategy. For this purpose, owing to the regression-based formulation of the OC-KSR method, first, it is shown that the kernel decomposition approach for learning vector-valued functions in the Hilbert space is directly applicable to the OC-KSR methodology which in turn facilitates learning a predictive one-class vector-valued function and a linear structure among multiple tasks, concurrently. Next, as a second contribution of the present work and in contrary to the common methods which model inter-target relations linearly in terms of an output matrix, a new \textit{non-linear} multi-task structure learning method is proposed where the relationship among multiple OCC problems is encoded via a non-linear kernel function. Task-specific coefficients, as well as output mixing parameters are then learned concurrently via a new alternating direction block minimisation method. Finally, it is shown that the proposed non-linear approach for one-class vector-valued function learning may be naturally extended to a sparse representation framework where different tasks compete in a sparse non-linear multi-task structure. To summarise, the main contributions of the current study may be outlined as:
\begin{itemize}
\item A separable kernel learning approach for multi-task Fisher null-space one-class classification where the structure among multiple problems is captured \textit{linearly} in terms of an output composition matrix;
\item A non-linear multi-task Fisher null-space one-class learning approach where the structure among multiple problems is modelled \textit{non-linearly} through a kernel function;
\item Extension of the non-linear multi-task structure learning mechanism to a sparse setting where the structure among multiple problems is encoded in a sparse fashion;
\item And, an experimental evaluation and comparison between different variants of the proposed multi-task one-class learning paradigm as well as other existing approaches on different datasets.
\end{itemize}
\subsection{Outline of the paper}
The rest of the paper is organised as follows: a summary of the existing work on the multi-task one-class learning problem as well as a brief overview of non-OCC multi-target regression approaches, most relevant to the current study, is provided in \S\ref{rw}. In \S \ref{back}, once an overview of the one-class kernel spectral regression method for one-class learning \cite{DBLP:journals/corr/abs-1807-01085,DBLP:journals/corr/abs-1902-02208} is presented, the vector-valued function learning methodology, with an emphasis on separable kernel learning in the RKHSvv, is briefly reviewed. The proposed multi-task one-class kernel null-space approach is introduced in \S \ref{MTOCKSR} where the linear and non-linear structure learning mechanisms subject to Tikhonov as well as sparse regularisation are presented. An experimental evaluation of the proposed structure learning methods on different datasets is carried out in \S \ref{exp} along with a comparison against the baseline as well as other existing approaches in the literature. Finally, \S \ref{conc} offers brief conclusions.
\section{Related Work}
\label{rw}
In this section, a brief overview of the existing multi-task one-class learning approaches is presented. A number of non-OCC multi-target regression methods, related to the present work, shall be briefly reviewed too. For a detailed review on the general concept of multi-task learning the reader is referred to \cite{DBLP:journals/corr/ZhangY17aa}.

As an instance of the multi-task learning approaches for OCC, in \cite{HE2014416}, based on the assumption of closeness of the related tasks and proximity of their corresponding models, two multi-task learning formulations based on one-class support vector machines are presented. Both multi-task learning methods are then solved by optimising the objective function of a single one-class SVM. Other work in \cite{7899861}, presents a multi-task approach to include additional new features into a one-class classification task. Based on the theory of support vector machines, a new multi-task learning approach is proposed to deal with the training of the updated system. In \cite{5596881}, based on the one-class $\nu$-SVM, an MTL framework for one-class classification is presented which constrains different problems to have similar solutions. Such a formulation is cast into a second-order cone programme to derive a global solution. In \cite{8215490}, the authors propose a method for anomaly detection when collectively monitoring many complex systems. The proposed multi-task learning approach is based on a sparse mixture of Gaussian graphical models (GGM's) where each task is represented by a mixture of GGM's providing the functionality to handle multi-modalities. A new regularised formulation is then proposed with guaranteed sparsity in mixture weights. By introducing the vector-valued function subject to regularisations in the vector-valued reproducing Hilbert kernel space, an unsupervised classifier to detect the outliers and inliers simultaneously is proposed in \cite{DBLP:conf/cikm/DangCWZC17} where preserving the local similarity of data in the input space is encouraged by manifold regularisation.

In the general context of multi-target regression and apart from the one-class classification paradigm, there exist a variety of different methods. These methods are not directly related to the present work as they do not solve a one-class classification problem. Nevertheless, similar to the present work, in these methods, the multi-task learning problem is formulated as one of kernel regression. As an instance, in \cite{ICML2011Dinuzzo_54}, an output kernel learning method based on the solution of a suitable regularisation problem over a reproducing kernel Hilbert space of vector-valued functions is proposed. A block-wise coordinate descent method is then derived that efficiently exploits the structure of the objective functional. Other work in \cite{Ciliberto:2015:CLM:3045118.3045283}, addresses the MTL problem by illustrating that multiple tasks and their structure can be efficiently learned by formulating the problem as a convex optimisation problem which is solved by means of a block coordinate method. More recently, in \cite{7888599}, a multi-target regression approach via robust low-rank learning is proposed. The proposed approach can encode inter-target correlations in a structure matrix by matrix elastic nets. Other method \cite{7879814} models intrinsic inter-target correlations and complex non-linear input-output relationships via multi-target sparse latent regression where inter-target correlations are captured via $L_{2,1}$-norm-based sparse learning. Other work \cite{LI2018134} presents a two layer approach to jointly learn the latent shared features among tasks and a multi-task model based on Gaussian processes. In \cite{JMLR:v17:15-602}, in order to take into account the structure in the input data while benefiting from kernels in the input space, the reproducing kernel Hilbert space theory for vector-valued functions is applied. In \cite{8513983}, the objective for multitask learning is formulated as a linear combination of two sets of eigenfunctions, such that the eigenfunctions for one task can provide additional information on another and help to improve its performance. For a detailed review on multi-target regression one may consult \cite{}.
\section{Background}
\label{back}
\subsection{One-Class Kernel Spectral Regression}
The Fisher criterion is a widely applied design objective in statistical pattern classification where a projection function from the input space into a feature space is inferred such that the between-class scatter of the data is maximised while minimising the within-class scatter:
\begin{eqnarray}
\boldsymbol{\varphi}^\star=\operatorname*{arg\,max}_{\boldsymbol{\varphi}}\frac{\boldsymbol{\varphi}^\top \mathbf{S}_b \boldsymbol{\varphi}}{\boldsymbol{\varphi}^\top \mathbf{S}_w \boldsymbol{\varphi}}
\label{null}
\end{eqnarray}
\noindent where $\mathbf{S}_b$ denotes the between-class scatter matrix, $\mathbf{S}_w$ stands for the within-class scatter matrix and $\boldsymbol{\varphi}$ is a basis defining one axis of the subspace. A theoretically optimal projection which provides the best separability with respect to the Fisher criterion is the \textit{null} projection \cite{DBLP:journals/corr/abs-1807-01085,6619277,8099922}, yielding a positive between-class scatter while providing a zero within-class scatter:
\begin{eqnarray}
\nonumber \boldsymbol{\varphi}^\top \mathbf{S}_w \boldsymbol{\varphi} = 0\\
\boldsymbol{\varphi}^\top \mathbf{S}_b \boldsymbol{\varphi} > 0
\label{nullF}
\end{eqnarray}
In a one-class classification problem, the single optimiser for Eq. \ref{null} is found as the eigenvector corresponding to the largest eigenvalue of the following eigen-problem:
\begin{eqnarray}
\mathbf{S}_b\boldsymbol{\varphi}=\lambda \mathbf{S}_w\boldsymbol{\varphi}
\label{geig}
\end{eqnarray}
Having determined the null projection direction, a sample $x$ is projected onto the null-space as
\begin{eqnarray}
\mathbf{y} = \boldsymbol{\varphi}^\top \mathbf{x}
\label{nullproj}
\end{eqnarray}
In order to handle data with inherently non-linear structure, kernel extensions of this methodology are proposed \cite{DBLP:journals/corr/abs-1807-01085,6619277,8099922}. While solving for the discriminant in a kernel space requires eigen-analysis of dense matrices, a computationally efficient method (one-class kernel spectral regression, a.k.a. OC-KSR) based on spectral regression is proposed in \cite{DBLP:journals/corr/abs-1807-01085} which poses the problem as one of solving a regularised regression problem in the Hilbert space:
\begin{eqnarray}
\mathbf{a}^{opt}=\operatorname*{arg\,min}_{\mathbf{a}}\left \{\Vert \mathbf{K}\mathbf{a}-\mathbf{r} \Vert_2^2+\gamma \mathbf{a}^\top\mathbf{K} \mathbf{a} \right\}
\label{reg}
\end{eqnarray}
\noindent where $\gamma$ is a regularisation parameter, $\mathbf{r}$ denotes the desired responses and $\mathbf{K}$ stands for the kernel matrix. The optimal solution $\mathbf{a}^{opt}$ to the problem above is given as
\begin{eqnarray}
\mathbf{a}^{opt} = (\mathbf{K}+\gamma \mathbf{I_n})^{-1}\mathbf{r}
\label{alphaeq}
\end{eqnarray}
\noindent where $\mathbf{I_n}$ denotes an identity matrix of size $n$ ($n$ being the number of training samples). Once $\mathbf{a}^{opt}$ is determined, the projections of samples onto the null feature space are found as $\mathbf{y}=\mathbf{K}\mathbf{a}^{opt}$. For classification, the distance between the projection of a test sample and that of the mean of the target class is employed as a dissimilarity criterion.

In a single-task OC-KSR approach, the procedure starts with building a separate kernel matrix for each one-class classification problem followed by assigning optimal responses for each individual observation in each task. The optimal response vector $\mathbf{r}$ in the OC-KSR algorithm when only positive instances are available for training is shown to be a vector of ones (up to a scale factor). When negative training observations are also available, they are mapped onto the origin \cite{}.
\subsection{Vector-valued functions in the Hilbert space}
\label{lvvf}
Let us assume there are $T$ scalar learning problems (tasks), each associated with a training set $D_t$ of $n_t$ input-output observations $D_t=\big \{(x^t_{i},r^t_{i})\big\}_{i=1}^{n_t}$ with $x^t_i\in \mathcal{X}$ input space and $r^t_i\in \mathbb{R}$ output space and $t\in \{1,\dots ,T\}$ indexing a task. Given a loss function $\mathcal{L}:\mathbb{R}\times\mathbb{R} \rightarrow \mathbb{R}_{+}$ that measures the per-task prediction errors, in the problem of learning vector-valued functions in the Hilbert space, one is interested in a vector-valued function $\mathbf{f}(.)$ which jointly minimises the regularised errors corresponding to multiple learning problems, i.e. $\mathbf{f}^*(.)=\operatorname*{arg\,min}_{\mathbf{f}\in \mathcal{H}}Q_{L}$, where $Q_{L}$ is defined as
\begin{eqnarray}
Q_{L}=\sum_{t=1}^{T}\frac{1}{n_t}\sum_{i=1}^{n_t}\mathcal{L}(r_i^t,f_t(x_i^t))+\mathcal{R}(\mathbf{f})
\label{rhksvv}
\end{eqnarray}
$\mathcal{R}(\mathbf{f})$ denotes a regularisation on the function $\mathbf{f}(.)$ with scalar components $f_t$ in the Hilbert space.
A popular sub-class of vector-valued function learning methods in the Hilbert space is that of multi-target kernel \textit{regression} problem where the loss function $\mathcal{L}$ encodes a least squares lost in the Hilbert space. A commonly applied simplifying assumption in this direction is the separability of input-output relations which leads to an expression of the function $\mathbf{f}(.)$ in terms of a separable kernel. Separable kernels are functions of the form $\boldsymbol{\Gamma}(x,x^\prime)=\kappa(x,x^\prime)\mathbf{B}$, where $\kappa:\mathcal{X}\times\mathcal{X}\rightarrow\mathbb{R}$ is a scalar reproducing kernel that captures similarities between the inputs and $\mathbf{B}$ is a symmetric positive semi-definite $T\times T$ matrix encoding dependencies among the outputs. In this case, $\mathbf{f}(.)$ is represented as
\begin{equation}
\mathbf{f}(.)=\sum_{i=1}^n\kappa(x_i,.)\mathbf{B}\mathbf{a}_i
\label{sep}
\end{equation}
where $\mathbf{a}_i$ stands for task-specific coefficients. The output on the training data shall then be derived as $\mathbf{KAB}$ and the regularised loss given in Eq. \ref{rhksvv} may be expressed in a matrix form as
\begin{eqnarray}
Q_L=\Vert \mathbf{KAB}-\mathbf{R}\Vert_2^2+\mathcal{R}(\mathbf{K,A,B})
\label{sepcost}
\end{eqnarray}
where $\mathbf{K}$ denotes the kernel matrix for the inputs, $\mathbf{A}^{n\times T}$ ($n=\sum_{t=1}^T n_t$) stands for a matrix of the coefficient vectors $\mathbf{a}_i$ and $\mathbf{R}$ denotes a matrix collection of the expected responses while $\Vert .\Vert_2^2$ denotes the Frobenius norm. For this class of kernels, if $\mathbf{B}$ is the identity matrix, all outputs would be treated as being unrelated and the solution to the multi-task problem will be similar to that of solving each task independently. When the output structure matrix $\mathbf{B}$ is presumed to be other than the identity matrix, the tasks are regarded as related and finding the optimal function $\mathbf{f}(.)$ is posed as learning the matrices $\mathbf{A}$ and $\mathbf{B}$, concurrently, subject to suitable regularisation constraints. The generic form of $Q_L$ expressed in Eq. \ref{sepcost} may be considered as the common formulation to the multi-target regression problem in the Hilbert space where the choice of the regularisation function $\mathcal{R}$ leads to different instantiations of the problem. With reference to the separable kernel learning formulation for multi-task learning, one may interpret the output as finding the intermediate responses corresponding to each individual task via $\mathbf{KA}$ (similar to the OC-KSR approach) and then mixing them via a structure encoding mechanism to produce the final responses. From this standpoint, the final responses may be considered as the output of a composition function $\mathbf{f}(.)=\mathbf{g}(\mathbf{h}(.))$, where $\mathbf{h}(.)$ produces the intermediate responses while $\mathbf{g}(.)$ performs a composition on the intermediate responses to derive the final output. From a compositional function perspective, the relations in Eqs. \ref{sep} and \ref{sepcost}, correspond to a non-linear mapping function $\mathbf{h}(.)$ expressed in terms of $\kappa(.,.)$ and $\mathbf{A}$, while the linear function $\mathbf{g}(.)$ is defined as a linear mixing function, characterised via matrix $\mathbf{B}$. The majority of the existing work on multi-task structure learning is focused on the case where $\mathbf{g}(.)$ is a linear function.
In this work, we study the problem of jointly learning multiple one-class classification problems by modelling individual task-predictors as the components of a vector-valued function.
In doing so, the utility of the compositional function view is demonstrated for learning structures among multiple OCC problems. For this purpose, we consider two general cases: 1-when the function $\mathbf{g}(.)$ is a linear function, we refer to the structure among multiple problems as a linear structure; and 2-when $\mathbf{g}(.)$ is a non-linear function, the structure would be referred to as a non-linear structure. Note that for both alternative scenarios above, function $\mathbf{h}(.)$ is assumed to be a non-linear function, defined in a Hilbert space. For a general Representer Theorem regarding compositional functions in the Hilbert space the reader is referred to \cite{Bohn.Griebel.Rieger:2017}.
\section{Multi-Task One-Class Kernel Null-Space}
\label{MTOCKSR}
In this section, first, the proposed multi-task one-class learning method for linear structure learning is introduced. The discussion is then followed by presenting a non-linear structure learning approach based on Tikhonov regularisation which is then modified to learn sparse non-linear multi-task structures. 
\subsection{Linear Structure Learning}
\label{lsl}
\begin{figure}[t]
\centering
\includegraphics[scale=.35]{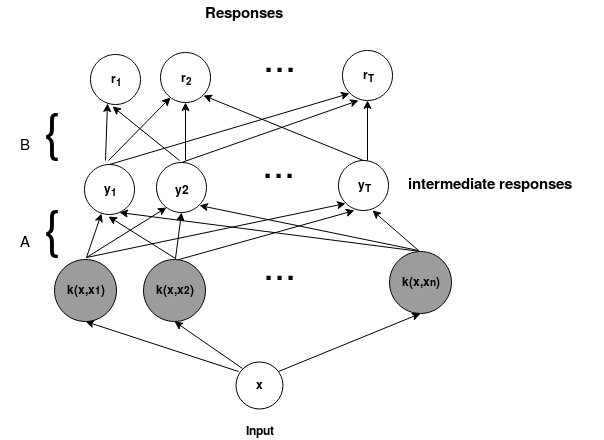}
\caption{Linear multi-task structure learning in the proposed multi-task one-class approach.}
\label{lin}
\end{figure}

In the linear set up of the proposed multi-task one-class learning method (i.e. when $\mathbf{g}(.)$ is a linear function), following the formulation in Eq. \ref{sepcost}, once the intermediate responses corresponding to different tasks are determined, they are mixed via an output matrix to produce the final responses. The key to the deployment of the cost function in Eq. \ref{sepcost} in the context of one-class classification based on the OC-KSR approach is that Eq. \ref{sepcost} is quite general with no restriction imposed on the responses $\mathbf{R}$. The only requirement for $Q_L$ to characterise a kernel null-space one-class classifier is that of suitable choices for the responses $\mathbf{R}$. For this purpose and in order to be consistent with the OC-KSR setting, a suitable choice for $\mathbf{R}$ is the one which forces all target observations to be mapped onto a single point distinct from the projection of any possible non-target samples. Choosing $\mathbf{R}$ as such would then lead to a zero within-class scatter while providing a positive between-class scatter, i.e. a null projection function. The learning machine induced by Eq. \ref{sepcost} admits a multi-layer structure where the second layer parameter $\mathbf{B}$ encodes a linear structure among multiple tasks whereas the first layer coefficients $\mathbf{A}$ represent a collection of task-specific parameters, Fig. \ref{lin}. The goal is then to concurrently learn the coefficient matrix $\mathbf{A}$ and the structure encoding matrix $\mathbf{B}$ subject to suitable regularisations on $\mathbf{A}$ and $\mathbf{B}$.
While there exists different methods characterised by different regularisations on the solution, recently, an effective approach is presented in \cite{7888599} to control the rank and shrinkage of $\mathbf{B}$ while penalising the norm of $\mathbf{A}$ in the Hilbert space. The advocated cost function in \cite{7888599} is defined as
\begin{eqnarray}
\nonumber Q_L=\Vert \mathbf{KAB}-\mathbf{R}\Vert_2^2+\gamma_{L1} \mbox{trace}(\mathbf{A^\top KA})\\
+\gamma_{L2} \mbox{trace}(\mathbf{B}^\top\mathbf{B})+\gamma_{L3} \mbox{trace}(\sqrt{\mathbf{B}^\top\mathbf{B}})
\end{eqnarray}
For the optimisation of the objective function, similar to other relevant approaches, a block coordinate descent method is suggested in \cite{7888599}, alternating between optimisation w.r.t. the parameters of the first layer and those of the second layer.
\subsubsection{Sub-problem w.r.t. $\mathbf{A}$}
The first block of variables for the minimisation of the objective function $Q_L$ is that of $\mathbf{A}$. In order to optimise $Q_L$ with respect to $\mathbf{A}$, the partial derivative of $Q_L$ with respect to $\mathbf{A}$ is set to zero:
\begin{eqnarray}
\frac{\partial Q_L}{\partial\mathbf{A}}=2\mathbf{K}(\mathbf{K}\mathbf{A}\mathbf{B}-\mathbf{R})\mathbf{B}+2\gamma_{L1}\mathbf{K}\mathbf{A}=0
\end{eqnarray}
A sufficient condition for the above equality to hold is
\begin{eqnarray}
\mathbf{K}\mathbf{A}\mathbf{B}\mathbf{B}-\mathbf{R}\mathbf{B}+\gamma_{L1} \mathbf{A}=0
\end{eqnarray}
The linear matrix equation above is known as the discrete-time Sylvester equation, commonly arisen in control
theory \cite{trove.nla.gov.au/work/21956847}. The solution to $\mathbf{A}$ is given as
\begin{eqnarray}
vec(\mathbf{A})=(\gamma_{L1}\mathbf{I} \otimes \mathbf{K}^{-1} + \mathbf{BB}\otimes \mathbf{I})^{-1}vec(\mathbf{K}^{-1}\mathbf{RB})
\label{Aeq}
\end{eqnarray}
where $\otimes$ stands for the Kronecker product and $vec(.)$ denotes a concatenation of the columns of a matrix into a vector. For large-scale problems, the solution above may be inefficient. In these cases, by utilising the structure of the problem, more efficient techniques have been devised \footnote{\label{myfootnote}\url{www.slicot.org}}.
\subsubsection{Sub-problem w.r.t. $\mathbf{B}$}
For the minimisation of the error function $Q_L$ with respect to $\mathbf{B}$, the work in \cite{7888599} proposed a gradient descent approach:
\begin{eqnarray}
\mathbf{B}= \mathbf{B}-\eta_B\partial Q_L/\partial\mathbf{B}
\label{Beq}
\end{eqnarray}
where $\eta$ is the step size parameter and $\partial Q_L/\partial\mathbf{B}$ is derived as
\begin{eqnarray}
\nonumber \frac{\partial Q_L}{\partial\mathbf{B}}=\frac{-2}{n}(\mathbf{KAB}-\mathbf{R})(\mathbf{AK})^\top\\+\beta\mathbf{U}\mathbf{\Sigma}^{-1}|\mathbf{\Sigma}|\mathbf{V}^\top
+2\gamma\mathbf{S}
\end{eqnarray}
where $\mathbf{B}=\mathbf{U}\mathbf{\Sigma}\mathbf{V}^\top$ is an eigen-decomposition of structure matrix and $|\mathbf{\Sigma}|$ is the matrix of element-wise absolute values of $\mathbf{\Sigma}$. For a detailed derivation of $\partial Q_L/\partial \mathbf{B}$ one may consult \cite{7888599}. An advantage of this approach over other alternatives lies in the convexity of the objective function which facilitates reaching the global optimum. Optimisation of the objective function $Q_L$ with respect to the unknown parameters $\mathbf{B}$ and $\mathbf{A}$ is then realised via an alternating direction minimisation approach, summarised in Algorithm \ref{locksr}, where during the initialisation step, all tasks are deemed to be independent. That is, the structure matrix $\mathbf{B}$ is initially set to identity.
A number of observations regarding the proposed one-class multi-task linear structure learning approach based on the OC-KSR method are in order. First, it should be noted that the structure in Fig. \ref{lin} depicts the \textit{learning} stage of the proposed one-class model. In the operational (test) phase, however, the parameter sets $\mathbf{A}$ and $\mathbf{B}$ may be combined to produce a model with a single layer of discriminants in the Hilbert space as $\mathbf{C}=\mathbf{A}\mathbf{B}$. As noted earlier, the structure considered in Fig. \ref{lin} is not new and has been previously explored in the context of multi-target regression. The novelty here, lies in posing the kernel null-space one-class classification approach in this context to benefit from the same learning structure, thanks to a kernel regression-based formulation of the OC-KSR approach.
\begin{algorithm}[t]
\caption{Linear multi-task OC-KSR}\label{locksr}
\begin{algorithmic}[1]
\State $\mathbf{B}=\mathbf{I}_T$
\Repeat
\State $vec(\mathbf{A})=(\gamma_{L1}\mathbf{I} \otimes \mathbf{K}^{-1} + \mathbf{BB}\otimes \mathbf{I})^{-1}vec(\mathbf{K}^{-1}\mathbf{RB})$
\State $\mathbf{B}= \mathbf{B}-\eta_B\partial Q_L/\partial\mathbf{B}$
\Until$|Q_L^{t+1}-Q_L^t|<\epsilon$
\end{algorithmic}
\normalsize
\end{algorithm}
\subsection{Non-linear Structure Learning}
In the proposed non-linear structure learning scheme and in contrast to the linear set-up, the relations among multiple tasks is modelled via a non-linear (kernel) approach. The structure of the learning machine proposed in this work for this purpose is illustrated in Fig. \ref{nl}. In this setting, once the intermediate responses corresponding to different tasks ($y_t$'s, for $t=1,\dots, T$) for a given input $\mathbf{x}$ are produced, they collectively serve as a single input ($\mathbf{y}$) to the second layer. In the second layer, $\mathbf{y}$ is then non-linearly mapped into a new space, induced by a kernel function (RBF kernel) and ultimately mixed via the coefficients $\mathbf{B}$ to produce the final responses corresponding to different tasks. The training data for the second layer thus consists of $T-$dimensional intermediate responses.
In the proposed non-linear structure learning method, the unknown matrices $\mathbf{A}$ and $\mathbf{B}$ are found by optimising an objective function $Q_{N}$ defined as a regularised kernel regression based on a kernel matrix $\mathbf{J}$ which captures the similarities between outputs of different tasks. The superiority of the non-linear model as compared with the conventional linear structure of Fig. \ref{lin} (as will be verified in the experimental evaluation section) may be justified from the perspective that the structure in Fig. \ref{lin} acts a linear regression of the intermediate responses while that of Fig. \ref{nl} corresponds to a non-linear (kernel) regression. Different regularisations in the proposed non-linear setting, namely Tikhonov and sparsity, are examined and discussed next.
\subsubsection{Tikhonov regularisation}
\begin{figure}
\centering
\includegraphics[scale=.35]{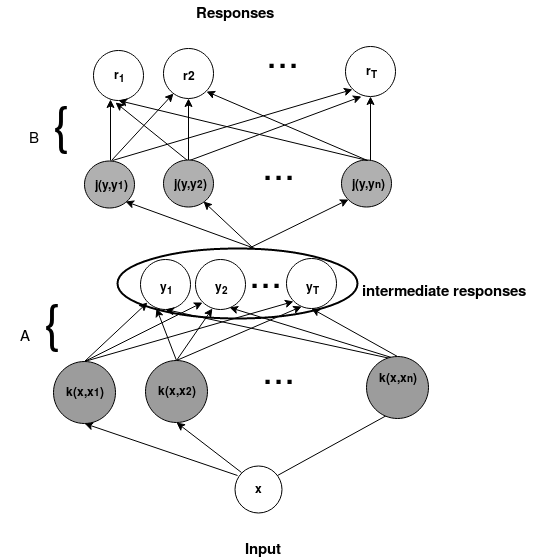}
\caption{Non-linear multi-task structure learning in the proposed multi-task one-class approach.}
\label{nl}
\end{figure}

A Tikhonov regularisation, in general, favours models that provide predictions that are as smooth functions of the intermediate responses as possible by penalising parameters of larger magnitude and thereby producing a more parsimonious solution. Following a Tikhonov regularised regression formulation in the Hilbert space, the objective function for the model in Fig. \ref{nl} is defined as
\begin{eqnarray}
\nonumber Q_{N}=\Vert \mathbf{J}\mathbf{B}-\mathbf{R}\Vert_2^2+\gamma_{N1}\mbox{trace}(\mathbf{A}^\top\mathbf{K}\mathbf{A})+\\
\gamma_{N2}\mbox{trace}(\mathbf{B}^\top\mathbf{J}\mathbf{B})
\label{qnl}
\end{eqnarray}
where $\mathbf{K}$ and $\mathbf{J}$ denote the kernel matrices associated with the first and the second layer, respectively. The optimisation of the objective function associated with the non-linear model is realised via a block coordinate descent scheme, alternating between optimisation w.r.t. the parameters of the first layer and those of the second layer.

\textit{Sub-problem w.r.t. $\mathbf{A}$}:
The first direction of minimisation for the objective function $Q_{N}$ is that of $\mathbf{A}$. The partial derivative of $\mbox{trace}(\mathbf{A}^\top \mathbf{K}\mathbf{A})$ w.r.t. $\mathbf{A}$ is readily obtained as
\begin{eqnarray}
\frac{\partial \mbox{trace}(\mathbf{A}^\top \mathbf{K}\mathbf{A})}{\partial\mathbf{A}}=2\mathbf{K}\mathbf{A}
\end{eqnarray}
Denoting the remaining terms of $Q_N$ as $Q_{N1}=\Vert \mathbf{J}\mathbf{B}-\mathbf{R}\Vert_2^2+\gamma_{N2}\mbox{trace}(\mathbf{B}^\top\mathbf{J}\mathbf{B})$, we shall proceed with computing its partial derivative w.r.t. $\mathbf{A}$. The parameters involved in $Q_{N1}$ are independent from $\mathbf{A}$ except for the kernel matrix $\mathbf{J}$. The dependency of the kernel matrix $\mathbf{J}$ on $\mathbf{A}$ is due to its dependency on the intermediate responses $\mathbf{Y}$ which is a function of $\mathbf{A}$ as $\mathbf{Y}=\mathbf{K}\mathbf{A}$. In order to compute the partial derivative of $Q_{N1}$ w.r.t. $\mathbf{A}$, first, the following matrices are defined:
\begin{eqnarray}
\nonumber \mathbf{F}&=&\mathbf{Y}\mathbf{Y}^\top\\
\nonumber \mathbf{E}&=&(\mathbf{I}\circ\mathbf{F})\mathbf{1}+\mathbf{1}^\top(\mathbf{I}\circ\mathbf{F})^\top-2\mathbf{F}\\
\end{eqnarray}
where $\circ$ stands for the Hadamard (component-wise) product and $\mathbf{1}$ denotes a matrix of ones. The kernel matrix $\mathbf{J}$ associated with the second layer may then be expressed as
\begin{eqnarray}
\mathbf{J}&=&\exp[-\theta \mathbf{E}]
\end{eqnarray}
where the scalar parameter $\theta$ controls the RBF kernel width associated with the second layer.
The partial derivative of $Q_{N1}$ with respect to the kernel matrix $\mathbf{J}$ is
\begin{eqnarray}
\frac{\partial Q_{N1}}{\partial\mathbf{J}}=2(\mathbf{J}\mathbf{B}-\mathbf{R})\mathbf{B}^\top+\gamma_{N2}\mathbf{B} \mathbf{B}^\top
\label{rondqj}
\end{eqnarray}
The partial derivatives $\partial Q_{N1}/\partial\mathbf{E}$, $\partial Q_{N1}/\partial\mathbf{F}$, $\partial Q_{N1}/\partial\mathbf{Y}$ are derived as \cite{10.1007/978-3-030-01216-8_38}
\begin{eqnarray}
\nonumber \frac{\partial Q_{N1}}{\partial\mathbf{E}}&=&(-\theta\mathbf{J})\circ \frac{\partial Q_{N1}}{\partial\mathbf{J}}\\
\nonumber \frac{\partial Q_{N1}}{\partial\mathbf{F}}&=&\mathbf{I}_n\circ\Big(\big(\frac{\partial Q_{N1}}{\partial\mathbf{E}} +{\frac{\partial Q_{N1}}{\partial\mathbf{E}}}^\top\big)\mathbf{1}^\top \Big)-2 \frac{\partial Q_{N1}}{\partial\mathbf{E}}\\
\frac{\partial Q_{N1}}{\partial\mathbf{Y}}&=& \big(\frac{\partial Q_{N1}}{\partial\mathbf{F}}+{\frac{\partial Q_{N1}}{\partial\mathbf{F}}}^\top\big)\mathbf{Y}
\end{eqnarray}
For the computation of $\partial Q_{N1}/\partial\mathbf{A}$, from the differentiation of a scalar-valued matrix function it is known that
\begin{eqnarray}
\delta Q_{N1}=\mbox{trace}({\frac{\partial Q_{N1}}{\partial \mathbf{Y}}}^\top\delta\mathbf{Y})=\mbox{trace}({\frac{\partial Q_{N1}}{\partial \mathbf{A}}}^\top\delta\mathbf{A})
\label{delta}
\end{eqnarray}
Since $\mathbf{Y}=\mathbf{K}\mathbf{A}$, it holds that $\delta\mathbf{Y}=\mathbf{K}\delta\mathbf{A}$. Replacing $\delta\mathbf{Y}$ by $\mathbf{K}\delta\mathbf{A}$ in Eq. \ref{delta} yields
\begin{eqnarray}
\delta Q_{N1}=\mbox{trace}({\frac{\partial Q_{N1}}{\partial \mathbf{Y}}}^\top\mathbf{K}\delta\mathbf{A})=\mbox{trace}({\frac{\partial Q_{N1}}{\partial \mathbf{A}}}^\top\delta\mathbf{A})
\end{eqnarray}
and hence
\begin{eqnarray}
\frac{\partial Q_{N1}}{\partial \mathbf{A}}=\mathbf{K}\frac{\partial Q_{N1}}{\partial \mathbf{Y}}
\end{eqnarray}
In summary, in order to compute $\partial Q_{N1}/\partial \mathbf{A}$, one first computes $\partial Q_{N1}/\partial \mathbf{J}$ and then $\partial Q_{N1}/\partial \mathbf{E}$, $\partial Q_{N}/\partial \mathbf{F}$ and $\partial Q_{N1}/\partial \mathbf{Y}$, respectively, followed by $\partial Q_{N1}/\partial \mathbf{A}$. Finally, $\partial Q_{N}/\partial \mathbf{A}=\partial Q_{N1}/\partial \mathbf{A}+2\gamma_{N1}\mathbf{K}\mathbf{A}$.

\textit{Sub-problem w.r.t. $\mathbf{B}$}:
Minimising the regularised error over multiple tasks represented by $Q_{N}$ w.r.t $\mathbf{B}$ may be performed by setting the partial derivative $\partial Q_{N}/\partial\mathbf{B}$ to zero:
\begin{eqnarray}
\nonumber \frac{\partial Q_{N}}{\partial\mathbf{B}}=2\mathbf{J}^\top(\mathbf{J}\mathbf{B}-\mathbf{R})+2\gamma_{N}\mathbf{J}\mathbf{B}=0
\end{eqnarray}
which yields
\begin{eqnarray}
\mathbf{B}=(\mathbf{J}+\gamma_{N}\mathbf{I}_n)^{-1}\mathbf{R}
\end{eqnarray}
Finally, the partial derivative of the objective function $Q_{N}$ with respect to $\theta$ is given as
\begin{eqnarray}
\frac{\partial Q_{N}}{\partial \theta}=\mbox{trace}\Big({\frac{\partial Q_{N}}{\partial \mathbf{J}}}^\top (-\mathbf{J}\circ\mathbf{E})\Big)
\end{eqnarray}
\textit{Initialisation}:
The initialisation step of the proposed non-linear structure learning model is similar in spirit to that of the linear case. That is, during the initialisation stage, each task is presumed as independent from all others. Based on this assumption, the kernel matrix $\mathbf{J}$ encoding inter-task relationships takes the form of a block-diagonal matrix, the diagonal elements of which are $\mathbf{1}^{n_t\times n_t}$ sub-matrices. Such an initialisation leads to an initialisation of $\mathbf{B}$ as $\mathbf{B}=(\mathbf{J}_{init}+\gamma_{N}\mathbf{I}_n)^{-1}\mathbf{R}$. The parameter controlling the width of the Gaussian kernel in the second layer ($\theta$) is initialised to the reciprocal of the average of $\mathbf{E}$. For the initialisation of $\mathbf{A}$, the problems are solved independently. Once all the parameters are initialised, the optimisation of the objective function with respect to the parameters of the first and the second layer is performed via an alternating direction minimisation scheme where for optimisation with respect to $\mathbf{A}$ and $\theta$ a gradient descent method is applied. The algorithm for the non-linear multi-task one-class learning is summarised in Algorithm \ref{nlocksr} where $\eta_A$ and $\eta_{\theta}$ denote the gradient descent step sizes for $\mathbf{A}$ and $\theta$, respectively. Note that in step 6 of the algorithm, the kernel matrix $\mathbf{J}$ associated with the second layer is updated based on the most recent values for $\mathbf{A}$ and $\theta$.
\begin{algorithm}[t]
\caption{Non-linear multi-task OC-KSR (Tikhonov regularisation)}\label{nlocksr}
\begin{algorithmic}[1]
\State $\mathbf{B}=(\mathbf{J}_{init}+\gamma_{N}\mathbf{I}_n)^{-1}\mathbf{R}$
\State $\theta=1/\mbox{mean}(\mathbf{E})$
\Repeat
\State $\mathbf{A}=\mathbf{A}-\eta_A \partial Q_{N}/\partial \mathbf{A}$
\State $\theta=\theta-\eta_{\theta} \partial Q_{N}/\partial \theta$
\State $\mathbf{J}=\mathbf{J}(\mathbf{A},\theta)$
\State $\mathbf{B}=(\mathbf{J}+\gamma_{N}\mathbf{I}_n)^{-1}\mathbf{R}$
\Until$|Q_N^{t+1}-Q_N^t|<\epsilon$
\end{algorithmic}
\normalsize
\end{algorithm}
\subsubsection{Sparse regularisation}
Apart from the widely used Tikhonov regularisation, other regularisation schemes encouraging sparseness on the solution are widely applied as a guideline for inference. The underlying motivation in this case is to provide the simplest possible explanation of an observation as a combination of as few as possible atoms from a given dictionary. A more compact model is expected to provide better performance as compared with its non-sparse counterpart, especially in the presence of corruption in data or missing relations between some problems. The sparsity in the proposed non-linear structure learning approach may be imposed at two different levels. The first level of sparsity is that of the task level: i.e. a task either contributes in forming the discriminant of another task (the tasks related) or not. The second level of sparsity is that of the within-task sparsity where the response for a particular problem is derived as a sparse representation of the corresponding training data. The two objectives above may be achieved via a sparse group lasso formulation \cite{slep,Yuan06modelselection} by enforcing an $L_{1}$-norm penalty on each element of $\mathbf{B}$ in addition to an $L_2$-norm task-wise penalty on $\mathbf{B}$. Consequently, the objective function $Q_{N}$ for the sparse non-linear setting is defined as 
\begin{eqnarray}
\nonumber Q_{N}=\Vert \mathbf{J}\mathbf{B}-\mathbf{R}\Vert_2^2+\gamma_{N1}\mbox{trace}(\mathbf{A}^\top\mathbf{K}\mathbf{A})\\+\gamma_{N2}\Vert\mathbf{B}\Vert_1^1+\gamma_{N3}\sum_{t=1}^T\Vert \mathbf{b}_t\Vert_2^2
\label{mtnls}
\end{eqnarray}
where $\gamma_{N2}$ controls the within-task sparsity while $\gamma_{N3}$ governs task-wise sparsity. As a result, in the proposed sparse multi-task one-class learning approach, each response $\mathbf{r}_t$ may be generated using only a few tasks from among the pool of multiple problems while at the same time using a sparse set of training observations.

The algorithm for the sparse non-linear multi-task one-class learning approach is similar to Algorithm \ref{nlocksr} except for two differences. First, when optimising w.r.t. $\mathbf{A}$, the partial derivative of $\partial Q_{N}/\partial \mathbf{J}$ would be
\begin{eqnarray}
\frac{\partial Q_{N}}{\partial\mathbf{J}}=2(\mathbf{J}\mathbf{B}-\mathbf{R})\mathbf{B}^\top
\label{rondqj2}
\end{eqnarray}
Second, in order to optimise the sparse group-lasso problem in Eq. \ref{mtnls} w.r.t. $\mathbf{B}$, the Sparse Learning with Efficient Projections (SLEP) algorithm \cite{slep} is used in this work. Using the SLEP algorithm, and by varying the regularisation parameters $\gamma_{N2}$ and $\gamma_{N3}$ solutions with different possible within-task or task-wise cardinalities on $\mathbf{B}$ may be obtained. The proposed sparse non-linear structure learning algorithm is summarised in Algorithm \ref{nlocksr2}.
\begin{algorithm}[t]
\caption{Non-linear multi-task OC-KSR (Sparse regularisation)}\label{nlocksr2}
\begin{algorithmic}[1]
\State $\mathbf{B}=(\mathbf{J}_{init}+\gamma_{N}\mathbf{I}_n)^{-1}\mathbf{R}$
\State $\theta=1/\mbox{mean}(\mathbf{E})$
\Repeat
\State $\mathbf{A}=\mathbf{A}-\eta_A \partial Q_{N}/\partial \mathbf{A}$
\State $\theta=\theta-\eta_{\theta} \partial Q_{N}/\partial \theta$
\State $\mathbf{J}=\mathbf{J}(\mathbf{A},\theta)$
\State $\mathbf{B}=\mbox{SLEP}(Q_{N})$
\Until$|Q_N^{t+1}-Q_N^t|<\epsilon$
\end{algorithmic}
\normalsize
\end{algorithm}
\subsection{Analysis of the algorithms}
A few comments regarding the dynamics of the proposed non-linear (Tikhonov/sparse) multi-task learning approaches are in order.

While in the linear structure learning method (Algorithm \ref{locksr}), the impact of changing one block of parameters on the other (the impact of $\mathbf{A}$ on $\mathbf{B}$ or vice versa) is explicit, in the non-linear setting (Algorithms \ref{nlocksr} and \ref{nlocksr2}) the two sets of parameters $\mathbf{A}$ and $\mathbf{B}$ interact indirectly via the kernel matrix associated with the second layer, i.e. via $\mathbf{J}$ (see step 6 of the Algorithms \ref{nlocksr} and \ref{nlocksr2}). Recall that the kernel matrix $\mathbf{J}$ associated with the second layer captures the similarities among multiple problems. In this respect, once $\mathbf{A}$ is updated, the intermediate responses are produced as $\mathbf{Y}=\mathbf{K}\mathbf{A}$. The new kernel matrix associated with the second layer may then be computed using the updated $\mathbf{Y}$ and the new parameter $\theta$. $\mathbf{B}$ is then derived based on the updated kernel matrix $\mathbf{J}$. Any modification to $\mathbf{B}$ would then affect parameter set $\mathbf{A}$ by making changes to $\partial Q_{N}/\partial \mathbf{J}$ (see Eqs. \ref{rondqj} and \ref{rondqj2}).

In the operation phase of the proposed non-linear structure learning methods, upon the arrival of a new test sample $\mathbf{x}$, the corresponding intermediate outputs ($y_t$'s for $t=1,\dots, T$) for different problems are produced by passing through the first layer. Treating the intermediate responses as the components of a single vector $\mathbf{y}=[y_1,\dots,y_T]^\top$, its similarity is measured to those of training samples associated with the second layer (i.e. $\mathbf{y}_i$'s for $i=1,\dots, n$) using a a kernel function and subsequently combined via the corresponding mixing matrix $\mathbf{B}$ to produce the final responses.
\section{Experimental Evaluation}
\label{exp}
In this section, an experimental evaluation of the proposed approaches for multi-task one-class classification is carried out.
\subsection{Data sets}
The efficacy of the proposed techniques is evaluated on three data sets, discussed next.
\subsubsection{Face} This data set is created to perform a toy experiment in face recognition. The data set contains face images of different individuals and the task is to recognise a subject among others. For each subject, a one-class classifier is built using the training data associated with the subject under consideration while all other subjects are assumed as outliers with respect to the model. The experiment is repeated in turn for each subject in the dataset. The features used for face image representation are obtained via the frontal-pose PAM deep CNN model \cite{8255649} applied to the face bounding boxes. The data set is created out of the real-access videos of the Replay-Mobile dataset \cite{Costa-Pazo_BIOSIG2016_2016} which is accompanied with face bounding boxes. In this work, ten subjects are used to form the data set. Each task is assumed to be the recognition of a single subject. The number of positive training samples for each subject is set to 4. The number of positive and negative test samples for each subject are 40 and 160, respectively where the negative test observations for each subject are selected randomly from subjects other than the subject under consideration.
\subsubsection{MNIST} MNIST is a collection of $28\times 28$ pixel images of handwritten digits 0-9 \cite{726791}. In our experiments, a single digit is considered as the target class while all others correspond to non-target observations. The experiment is repeated in turn for all digits. Similar to the face data set, each task is assumed to be the recognition of a digit among others. The number of positive training samples for each digit is set to 15. The number of positive and negative test samples for each class are set to 150 and 1350, respectively.
\subsubsection{Coil-100} The Coil-100 data set \cite{coil} contains 7,200 images of 100 different objects. Each object has 72 images taken at pose intervals of 5 degrees, with the images being of size $32\times 32$ pixels. In the experiments conducted on this data set, 50 classes are selected randomly. A one-class classifier is then trained to recognise an object of interest from others and considered as a single task. Raw pixel intensities are used as feature representations in this data set. The number of positive train instances for each target class is 7. 65 positive and 585 negative test observations for each class are included in the experiments on this data set.
\begin{figure*}[t]
\centering
\includegraphics[scale=.2]{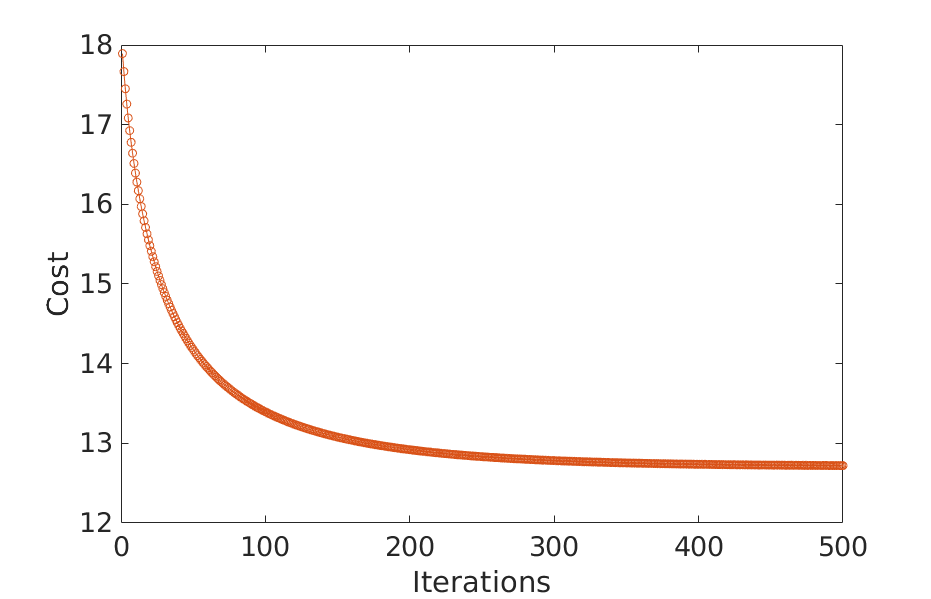}
\includegraphics[scale=.2]{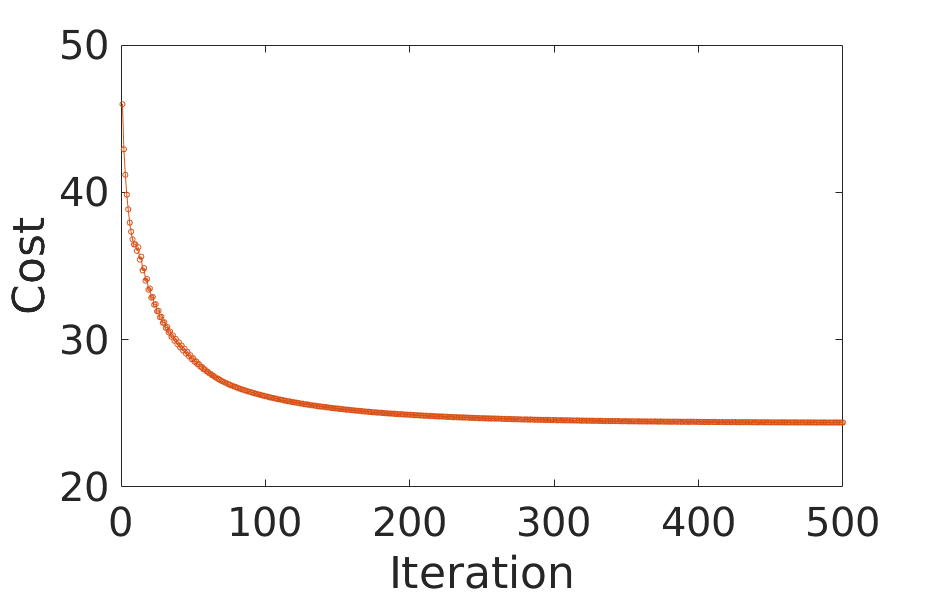}
\includegraphics[scale=.2]{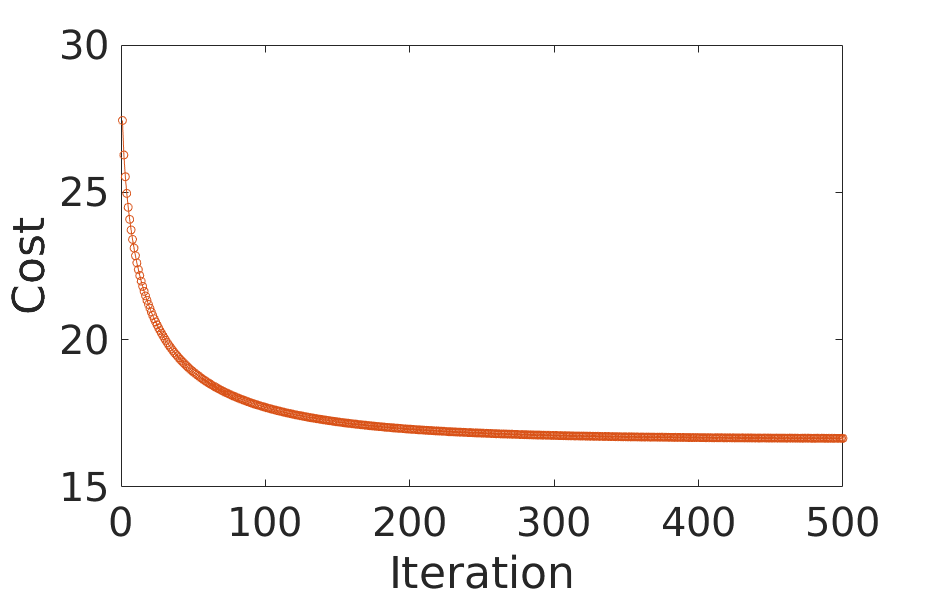}
\caption{Sample optimisation curves for the non-linear structure learning approach. From left to right: the face, the MNIST and the Coil-100 data sets.}
\label{opt_curves1}
\end{figure*}
\begin{figure*}[t]
\centering
\includegraphics[scale=.2]{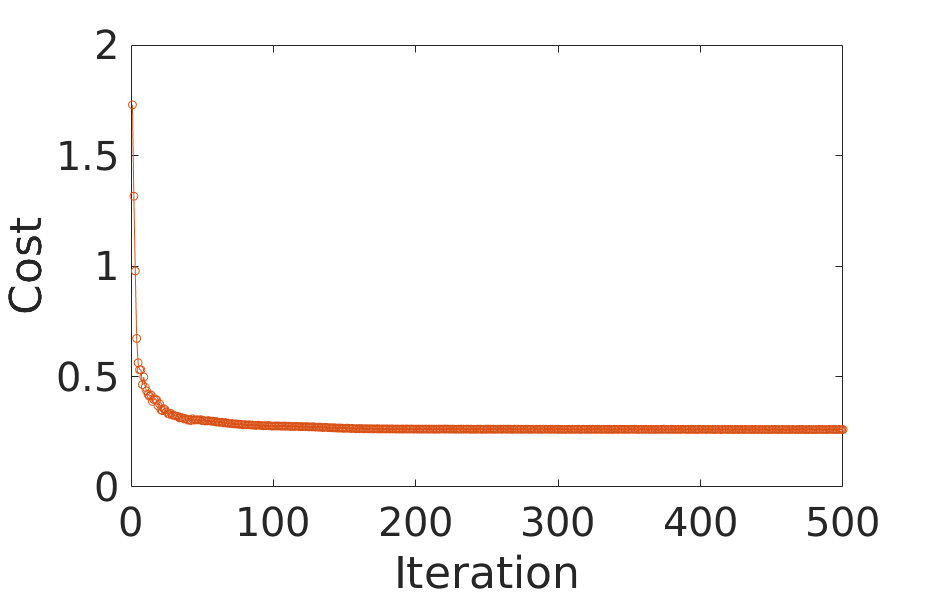}
\includegraphics[scale=.2]{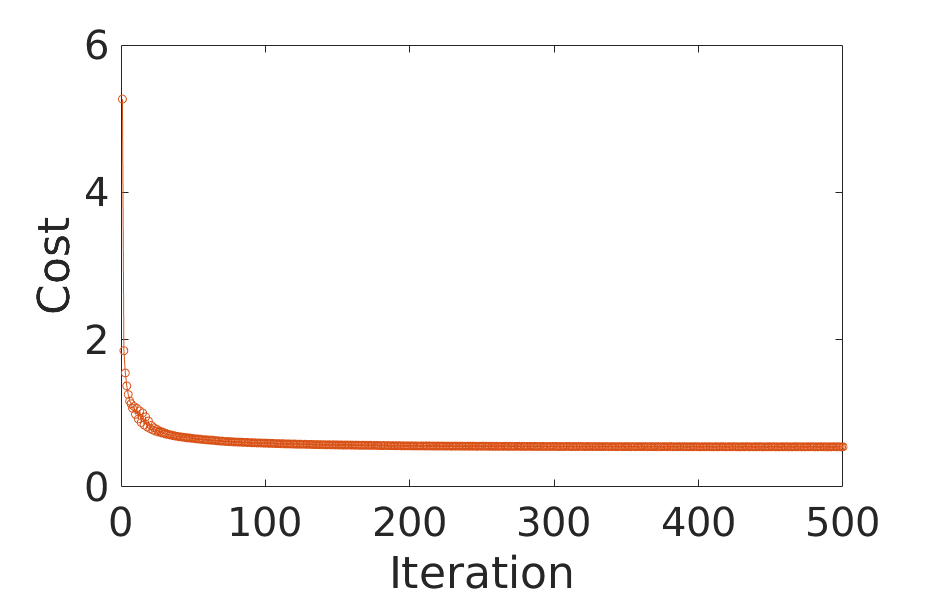}
\includegraphics[scale=.2]{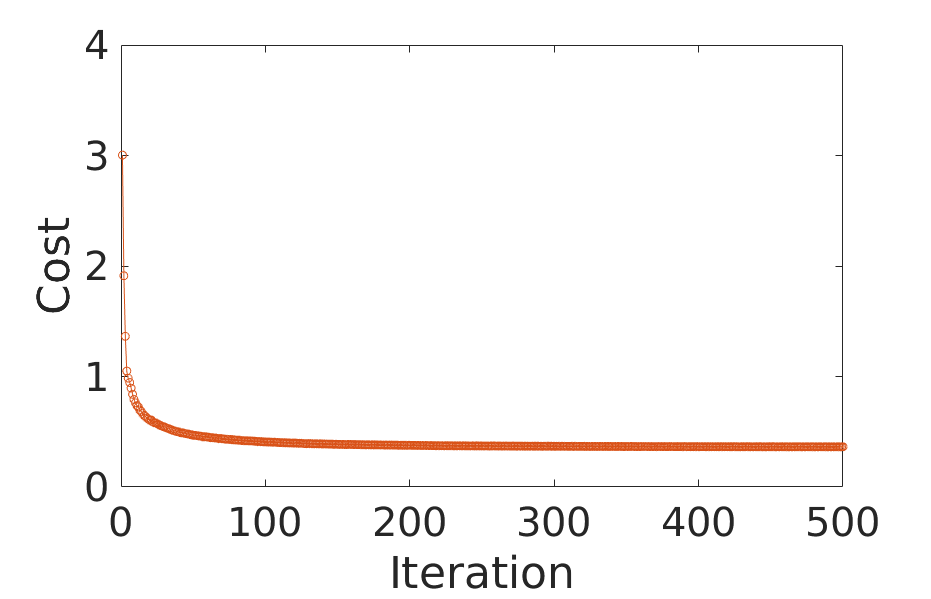}
\caption{Sample optimisation curves for the non-linear sparse structure learning approach. From left to right: the face, the MNIST and the Coil-100 data sets.}
\label{opt_curves2}
\end{figure*}
\begin{figure}[t]
\centering
\includegraphics[scale=5]{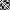}
\includegraphics[scale=5]{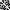}
\includegraphics[scale=1]{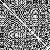}
\caption{Sample structural matrices for the linear structure learning approach. From left to right: the face, the MNIST and the Coil-100 data sets.}
\label{L_S}
\end{figure}
\begin{figure}[t]
\centering
\includegraphics[scale=1]{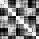}
\includegraphics[scale=.25]{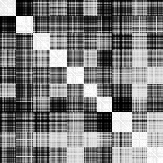}
\includegraphics[scale=.12]{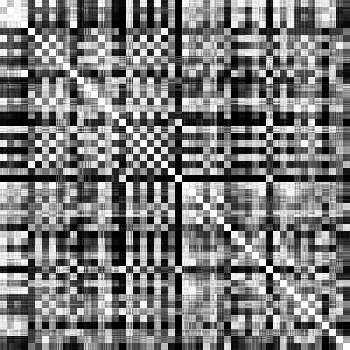}
\caption{Sample kernels for the non-linear structure learning approach. From left to right: the face, the MNIST and the Coil-100 data sets.}
\label{N_S}
\end{figure}
\begin{figure}[t]
\centering
\includegraphics[scale=1]{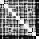}
\includegraphics[scale=.25]{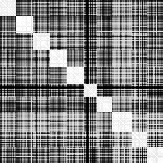}
\includegraphics[scale=.12]{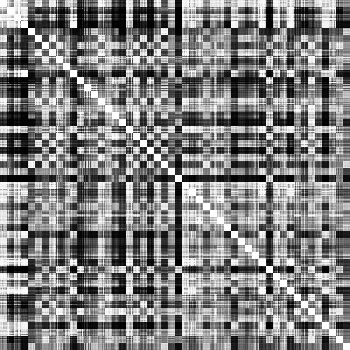}
\caption{Sample kernels for the non-linear sparse structure learning approach. From left to right: the face, the MNIST and the Coil-100 data sets.}
\label{NS_S}
\end{figure}
\subsection{Methods}
For the conversion of the OC-KSR method from a single-task to a multi-task setting, the first step is that of combining all training observations and forming a joint kernel matrix. In this case, the positive instances of one problem would serve as negative observations for all the remaining tasks. The optimal response would then be an $n\times T$ matrix where each row of the matrix is a vector of zeros except for a single element of one corresponding to the true class of the observation. 
As previously demonstrated in \cite{DBLP:journals/corr/abs-1807-01085}, utilisation of any possible negative training samples may boost the performance of the OC-KSR approach. In order to make a distinction between different variants of the OCKSR methodology, in this section, OCKSR would correspond to the algorithm when negative instances are not used for training while C-OCKSR shall be used to refer to the case when both positive and negative samples are used for training. This distinction is necessary to accurately gauge the benefits offered by a multi-task learning scheme independent of the effects of using non-target samples for training.
A thorough evaluation and comparison between different one-class classification algorithms has been conducted in \cite{DBLP:journals/corr/abs-1807-01085} and \cite{DBLP:journals/corr/abs-1902-02208} with the outcome of the OCKSR approach performing the best among other competitors. As such, different methods included in these experiments are:
\begin{itemize}
\item OCKSR is the original single-task OC-KSR method presented in \cite{DBLP:journals/corr/abs-1807-01085}. This method is used to learn an OCC classifier independently for each task and will serve as a baseline.
\item C-OCKSR corresponds to the single-task OCKSR approach where negative observations are utilised for training.
\item OCKSR-L is the proposed multi-task OCKSR approach where a linear structure between different tasks is learned.
\item OCKSR-N is the proposed multi-task OCKSR approach where a non-linear structure between different tasks subject to Tikhonov regularisation is learned.
\item OCKR-NS is the proposed multi-task OCKSR approach where a non-linear structure between different tasks subject to sparse group regularisation is learned.
\item SVDD is the Support Vector Data Description approach to solve the one class classification problem \cite{Tax2004}. As a widely used method, this method is used to learn an OCC classifier independently for each task to serves as a second baseline for comparison.
\item MORVR is the multi-output relevance vector regression \cite{DBLP:journals/corr/Ha17} which uses the Bayes’ theorem and the kernel trick to perform regression. The algorithm t uses the matrix normal distribution to model correlated outputs.
\end{itemize}
\subsection{Behaviour of the Optimisation Algorithms}
In this section, the effectiveness of the proposed alternating direction minimisation scheme for the non-linear setting (for both Tikhonov and sparse regularisation) is visualised. For an analysis of the convergence behaviour of the linear structure learning method one may consult \cite{7888599}. The optimisation curves depicting the cost function vs. iterations for the Tikhonov and sparse regularisation are illustrated in Fig. \ref{opt_curves1} and \ref{opt_curves2}, respectively. From Figs. \ref{opt_curves1} and \ref{opt_curves2}, one may observe that the proposed alternating direction approach converges within a few hundred iterations, irrespective of the nature of the observations. Interestingly, the convergence of the sparse approach seems to be relatively faster than its non-sparse counterpart. 
\subsection{Visualisation of Structure matrices}
The structures learned using different linear and non-linear multi-task approaches are illustrated in Figs. \ref{L_S}, \ref{N_S} and \ref{NS_S}, for the linear, non-linear and sparse non-linear setting, respectively. For the linear learning scheme, matrix $\mathbf{B}$ is visualised while for the non-linear setting, the kernel matrix associated with the second layer ($\mathbf{J}$) is illustrated. Note that for the Coil-100 data set, as a relative larger number of training sample are used, the kernel matrix is bigger in dimension compared to the other two data sets. In the figure, the kernel matrix for this data sets is rescaled to a similar size as those of others for visualisation purposes. As noted earlier, for initialisation, the structural matrices are set to (block)-diagonal matrices. As may be observed from the figures, for all data sets, the linear and non-linear multi-task learning approaches are successful to recover inter-task relations. This may be verified as all structural matrices incorporate non-zero off (block)-diagonal elements.
\begin{table*}
\small
\renewcommand{\arraystretch}{1.2}
\caption{Average performance (in terms of AUC ($\%$)) of different methods in a one-class classification scenario on different data sets}
\label{Characteristics}
\centering
\begin{tabular}{lccccccccc}
\hline
\textbf{Method} & \textbf{OCKSR} & \textbf{C-OCKSR} & \textbf{OCKSR-L} & \textbf{OCKSR-N} & \textbf{OCKSR-NS} & \textbf{SVDD} & \textbf{MORVR}\\
\hline
Face &97.70&99.55&99.71&99.78&99.76&97.69&97.63\\
MNIST &89.55&96.91&97.23&97.74&97.39&89.51&95.43\\
Coil-100 &92.08&97.32&97.40&98.87&97.95&93.18&78.27\\
\hline
\label{AUCTab}
\end{tabular}
\end{table*}

\begin{figure*}
\centering
\includegraphics[scale=.22]{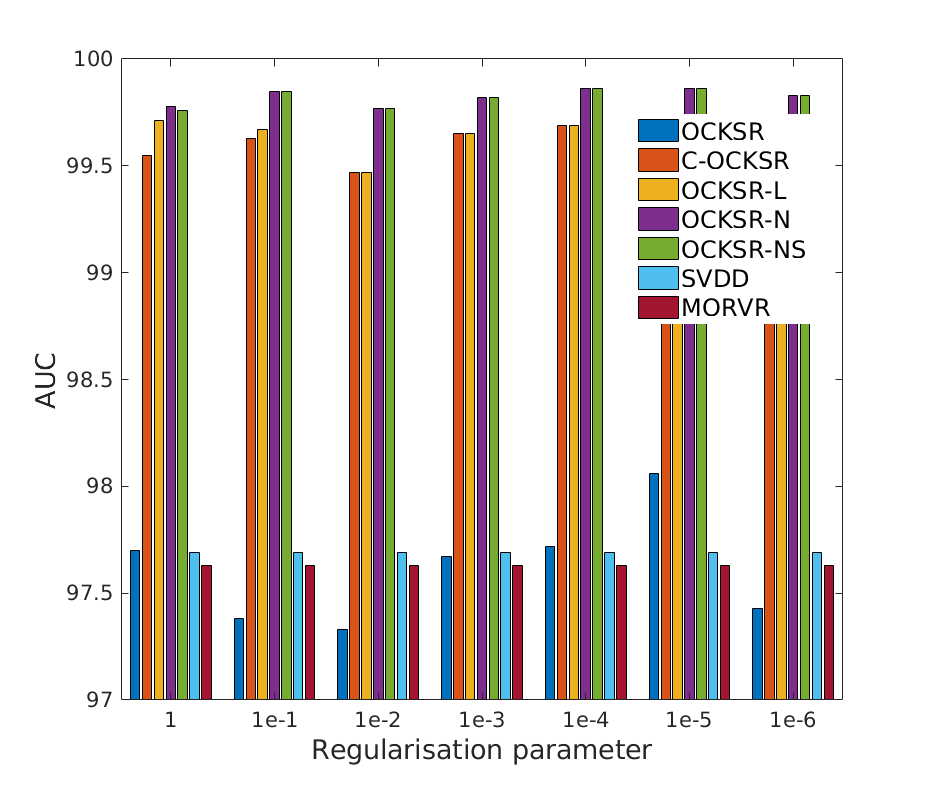}
\includegraphics[scale=.22]{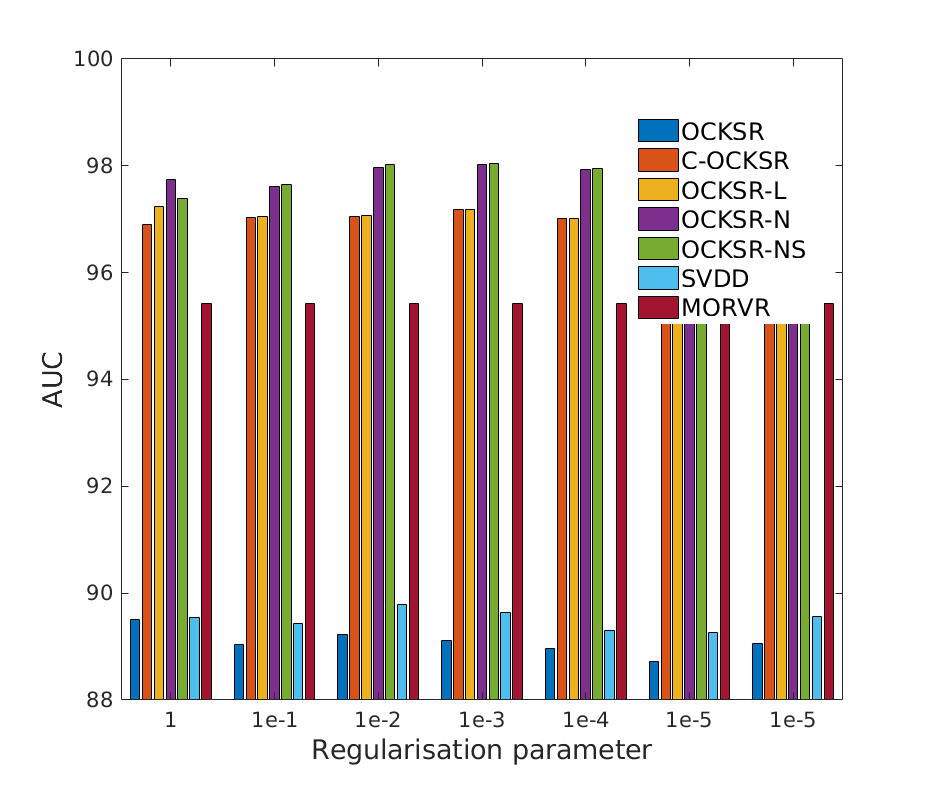}
\includegraphics[scale=.22]{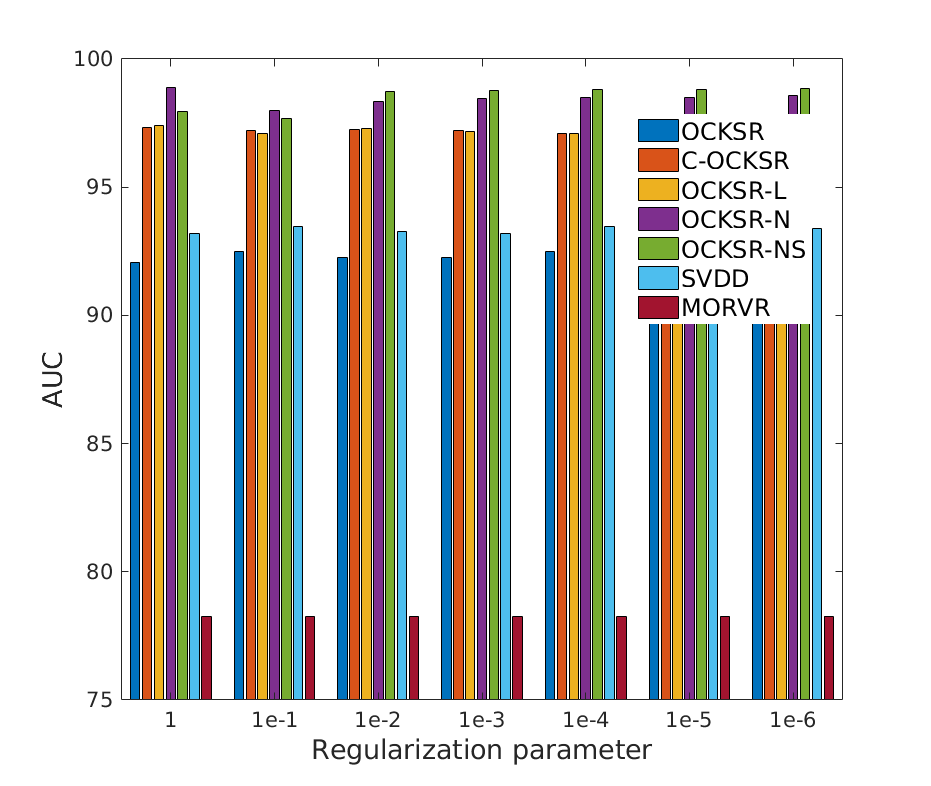}
\caption{The effect of regularisation on performance. From left to right: the face, MNIST and Coil-100 data sets.}
\label{reg}
\end{figure*}

\begin{figure*}
\centering
\includegraphics[scale=.2]{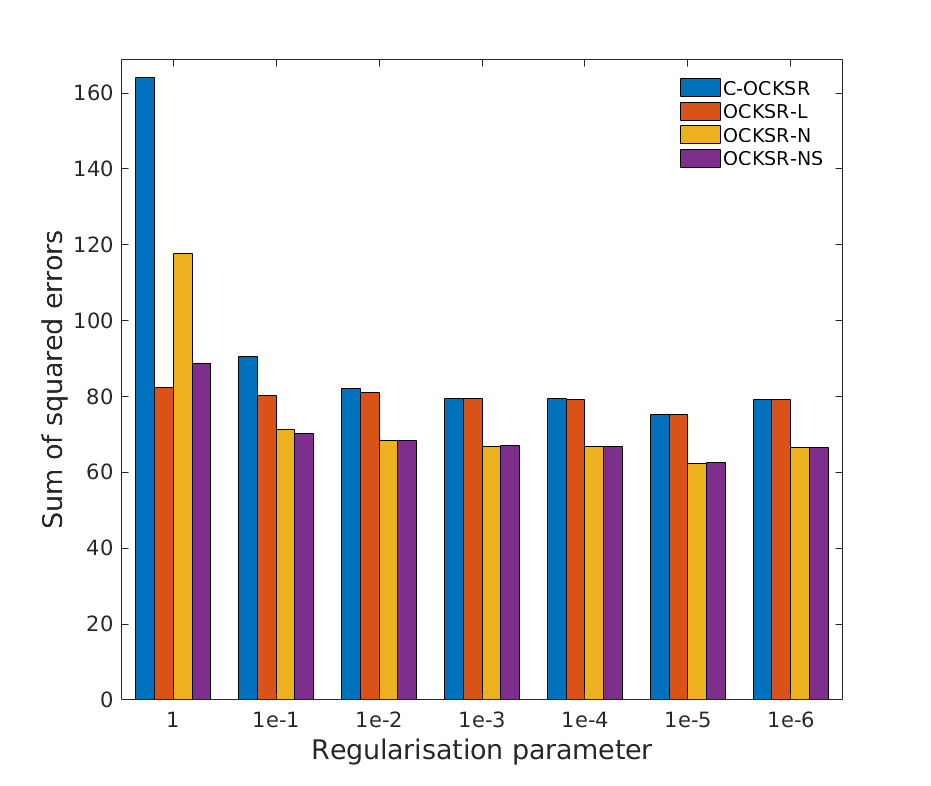}
\includegraphics[scale=.25]{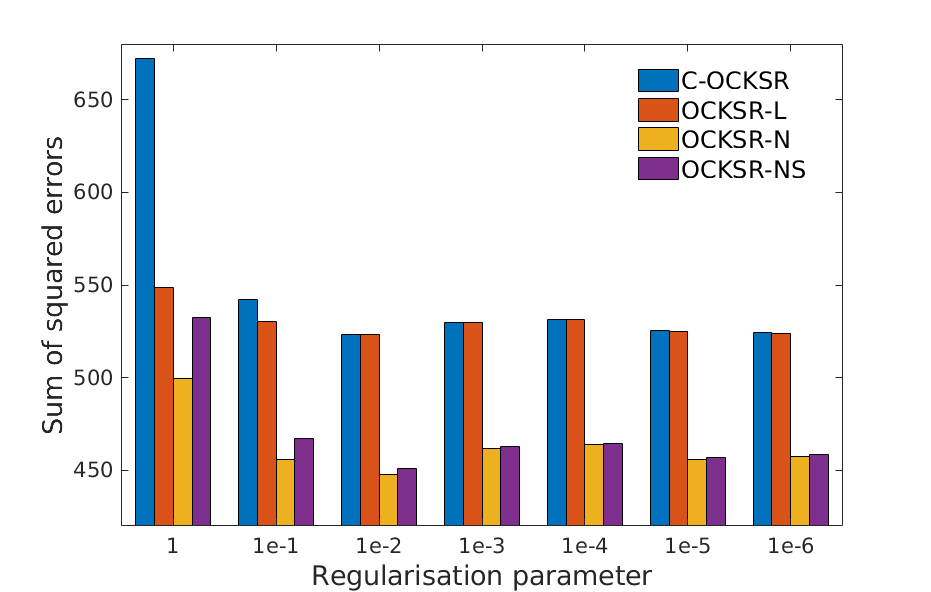}
\includegraphics[scale=.25]{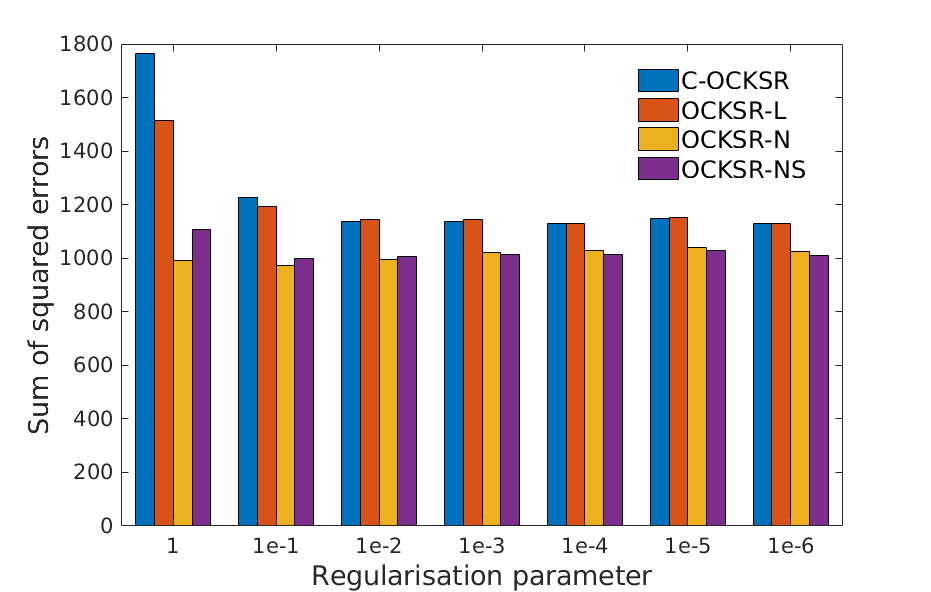}
\caption{The effect of regularisation on the sum of squared errors in predictions. From left to right: the face, MNIST and Coil-100 data sets.}
\label{rec}
\end{figure*}
\subsection{Performance Comparison}
In order to gauge the efficacy of the proposed multi-task OCC learning methods, multiple experiments are conducted on the face, MNIST and Coil-100 datasets subject to random partitions of the data into the train and test sets in order to minimise any bias associated with the partitioning of the data. For this purpose, each experiment is repeated 10 times and the average AUC measures are reported in Table \ref{AUCTab}. A number of observations from Table \ref{AUCTab} are in order. First, the proposed multi-task Fisher null-space approaches are effective in improving the performance compared to the single task OCKSR and the C-OCKSR methods. Second, from among the proposed multi-task learning scheme, the non-linear learning methods perform better than their linear counter-part which demonstrates the effectiveness of the proposed non-linear multi-task structure learning mechanism. Third, the proposed OCKSR-N method performs slightly better than its sparse counterpart. Nevertheless, the sparse method may provide an edge over the non-sparse variant when the data is corrupted.
\subsection{The Effect of regularisation}
In the experiments conducted thus far, the regularisation parameter corresponding to the first layer is set to 1 for all methods while the other parameters were optimised on the training set via cross validation. Typically a stronger regularisation reduces the flexibility of the model but may provide relatively more robustness against data corruption. In a final set of experiments the effect of changing the regularisation parameter of the first layer of the methods is analysed. For this purpose, the first layer regularisation parameter is chosen from $\{1,1e-1,1e-2,1e-3,1e-4,1e-5,1e-6\}$. The performances of different methods in terms of AUC are reported in Fig. \ref{reg}. Plots corresponding to sum of squared errors corresponding to different variants of the OCKSR method are provided in Fig. \ref{rec}. Note that since the error measures corresponding to other remaining methods were higher than the OCKSR variants, they are excluded from the figure in order to better analyse the effects of multi-target learning schemes. A number of observations from the figures are in order. First, the proposed non-linear structure learning methods perform better than other alternatives irrespective of the degree of regularisation, justifying the efficacy of a non-linear structure learning mechanism. Second, while the linear structure learning approach OCKSR-L provides an edge over the single-task C-OCKSR method for stronger regularisation, the advantage of learning a linear structure among multiple one-class problems vanishes towards lower regularisations where the performances of the methods nearly match. This may be observed from both the AUC as well as the error plots. Third, although the proposed sparse non-linear structure learning approach performs slightly inferior as compared with the non-sparse alternative for stronger levels of regularisation, nevertheless, towards lower regularisation levels it performs better than the non-sparse counterpart. Similar behaviour is observed both in terms of the AUC as well as the sum of squared error measure. 

\section{Conclusion}
\label{conc}
We have studied one-class classification based on the kernel Fisher null-space technique (OCKSR) in a multi-task learning framework. For this purpose, first, it was shown that the OCKSR approach may be readily cast within a multi-target learning approach where the dependencies among multiple tasks are modelled linearly. Next, a non-linear structure learning mechanism was proposed where the correlations among different problems were encoded more effectively. The non-linear multi-task learning approach was then extended to a sparse setting to account for any missing relationships among different problems. The experiments verified the merits of multi-task learning for the OCC problem based on OCKSR. Moreover, while in certain cases the common linear structure learning approach failed to provide advantages, the proposed non-linear multi-task learning methods maintained their edge over other alternatives.

\section*{Acknowledgment}
The authors would like to thank...
\ifCLASSOPTIONcaptionsoff
  \newpage
\fi
\bibliographystyle{IEEEtran}
\bibliography{IEEEexample2}
%
%
\begin{IEEEbiography}[{\includegraphics[width=1in,height=1.25in,clip,keepaspectratio]{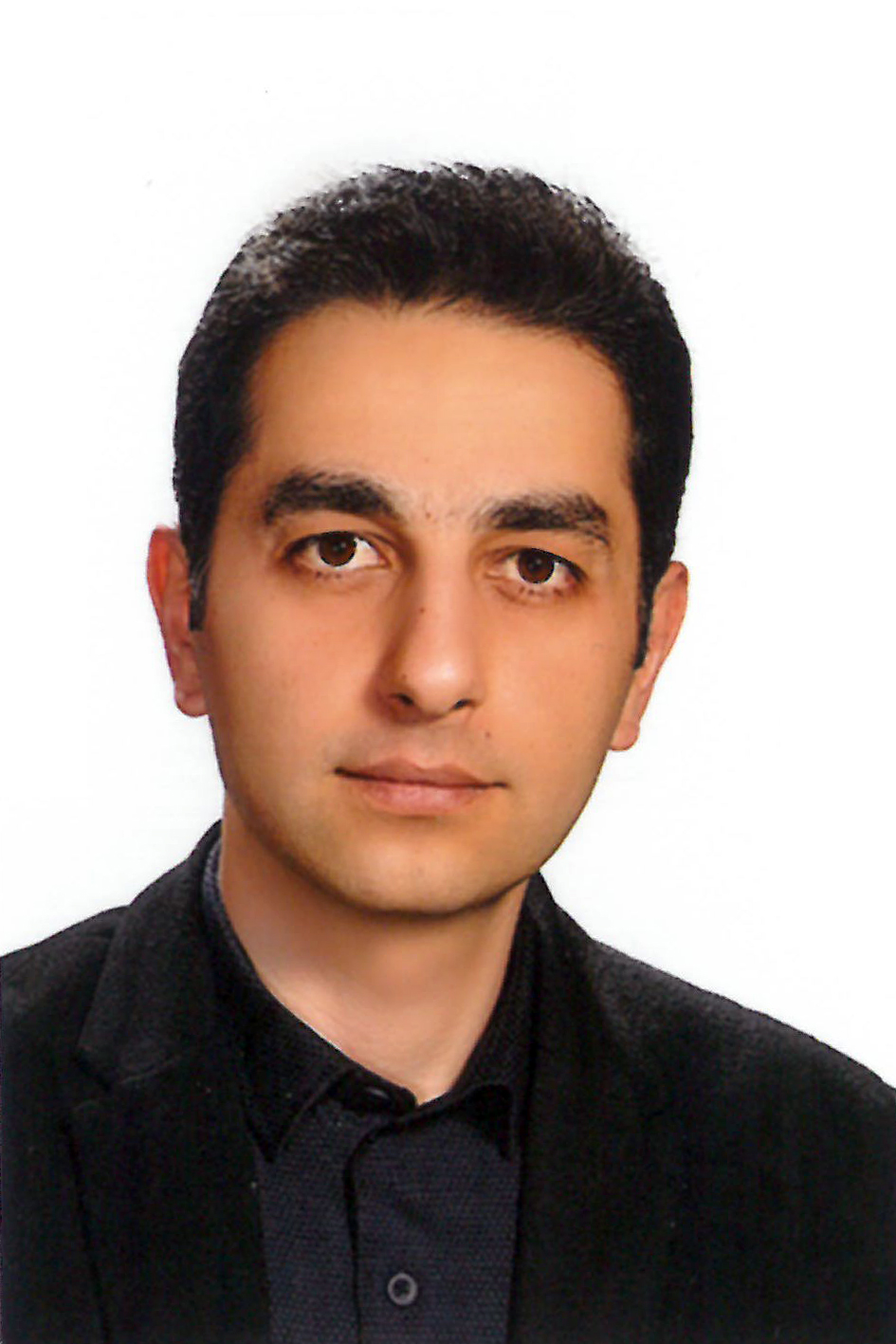}}]{Shervin Rahimzadeh Arashloo}
received the Ph.D. degree from the centre for vision, speech and signal processing, university of Surrey, UK. He is an assistant professor with the Department of Computer Engineering, Bilkent University, Ankara, Turkey and also holds a visiting research fellow position with the centre for vision, speech and signal processing, university of Surrey, UK. His research interests includes secured biometrics, anomaly detection and graphical models with applications to image and video analysis.
\end{IEEEbiography}

\begin{IEEEbiography}[{\includegraphics[width=1in,height=1.25in,clip,keepaspectratio]{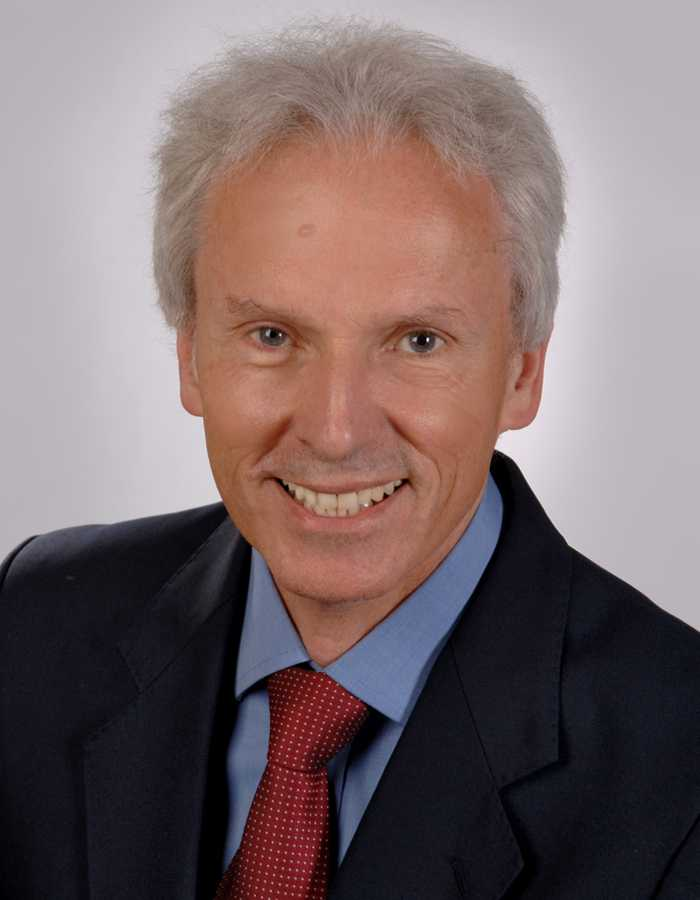}}]{Josef Kittler}
received the B.A., Ph.D., and D.Sc. degrees from the University of Cambridge, in 1971, 1974, and 1991, respectively. He is Professor of Machine Intelligence at the Centre for Vision, Speech and Signal Processing, Department of Electronic Engineering, University of Surrey, Guildford, U.K. He conducts research in biometrics, video and image database retrieval, medical image analysis, and cognitive vision. He published the textbook Pattern Recognition: A Statistical Approach (Englewood Cliffs, NJ, USA: Prentice-Hall, 1982) and over 600 scientific papers. He serves on the Editorial Board of several scientific journals in pattern recognition and computer vision.
\end{IEEEbiography}
\end{document}